%% file: main.tex
\documentclass[runningheads]{llncs}

\usepackage{eccv}

\usepackage{eccvabbrv}

\usepackage{graphicx}
\usepackage{booktabs}

\usepackage[accsupp]{axessibility}  

\usepackage{hyperref}

\usepackage{orcidlink}

\begin{document}

\title{Pushing Joint Image Denoising and Classification to the Edge} 


\author{Thomas C. Markhorst*\orcidlink{0009-0001-1597-6140} \and
Jan C. Van Gemert\orcidlink{0000-0002-3913-2786} \and
Osman S. Kayhan\orcidlink{0000-0002-0328-7647}}

\authorrunning{T.C. Markhorst et al.}

\institute{CV Lab, Delft University of Technology 
\\
\email{\{t.c.markhorst\}@tudelft.nl}}

\maketitle

\input{sec/0_abstract}    
\input{sec/1_intro}

\input{sec/2_related_work}

\input{sec/3_toy_setup}
\input{sec/4_nas}    
\input{sec/5_conclusion}

%
%
\bibliographystyle{splncs04}
\bibliography{main}

\input{sec/X_suppl}

\end{document}

%% file: sec/0_abstract.tex
\begin{abstract}
   In this paper, we jointly combine image classification and image denoising, aiming to enhance human perception of noisy images captured by edge devices, like low-light security cameras. In such settings, it is important to retain the ability of humans to verify the automatic classification decision and thus jointly denoise the image to enhance human perception.   
   Since edge devices have little computational power, we explicitly optimize for efficiency by proposing a novel architecture that integrates the two tasks. Additionally, we alter a Neural Architecture Search (NAS) method, which searches for classifiers to search for the integrated model while optimizing for a target latency, classification accuracy, and denoising performance. The NAS architectures outperform our manually designed alternatives in both denoising and classification, offering a significant improvement to human perception. Our approach empowers users to construct architectures tailored to domains like medical imaging, surveillance systems, and industrial inspections.
\end{abstract}

%% file: sec/1_intro.tex
\label{sec:intro}
\section{Introduction}
The intersection of edge devices, such as security cameras, and deep learning has sparked an interest in optimizing neural networks for inference time, further referred to as latency. Common tasks to optimize for such efficiency are object classification and object detection, which unlock automatic recognition. However, in noisy settings, the recognition accuracy might not be perfect and it is important to allow the ability to validate the automatic recognition by human inspection. Thus, in addition to automatic recognition, the perceptual quality of the processed image is equally significant. In particular, this is relevant for images containing noise, which can arise from various sources such as low-light conditions, sensor noise, or other recording conditions. We focus on using an efficient model that can be used on the edge with the aim of enhancing human perception for validating the recognition output of noisy images.

Domains relying on human image perception but challenged by noisy images, like medical imaging \cite{Medical_application}, surveillance systems \cite{Suveillance_application}, and industrial inspections \cite{Industrial_application}, can benefit from recently proposed denoising Convolutional Neural Networks (CNNs) \cite{DnCNN, RDUNet}. As CNNs denoise better than traditional methods \cite{BM3D, WNNM}. Fast CNN denoisers \cite{FFDNet, SGN} are required to accommodate the real-time requirement of the affected domains. However, denoisers are not able to remove all noise, which is not always enough for human image perception.

We further improve human understanding of the image by combining denoising with machine perception, like image classification. From the Human-Computer Cooperation strategies in \cite{HumanClassifierCoop}, we use the Classifier as Aid to Human. Where the image classifier can be used as a direct advisor or an independent agent to the security guard, the latter being analogous to a second opinion in medical diagnosis. In different use cases, fusing skills of humans and computers has been shown to improve performance beyond using only one type \cite{HumanClassifierCoop, HCI_ZoneOuts, HCI_underwater}. Therefore, we investigate models that can leverage the benefits of both denoising and classification to enhance human understanding in real time.

\begin{figure}[t]
  \centering
  \begin{minipage}{0.16\columnwidth}
    \centering
    \includegraphics[width=0.8\linewidth]{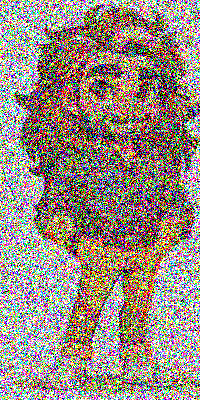}
    \subcaption{}
    \label{fig:subfigA}
  \end{minipage}%
  \begin{minipage}{0.16\columnwidth}
    \centering
    \includegraphics[width=0.8\linewidth]{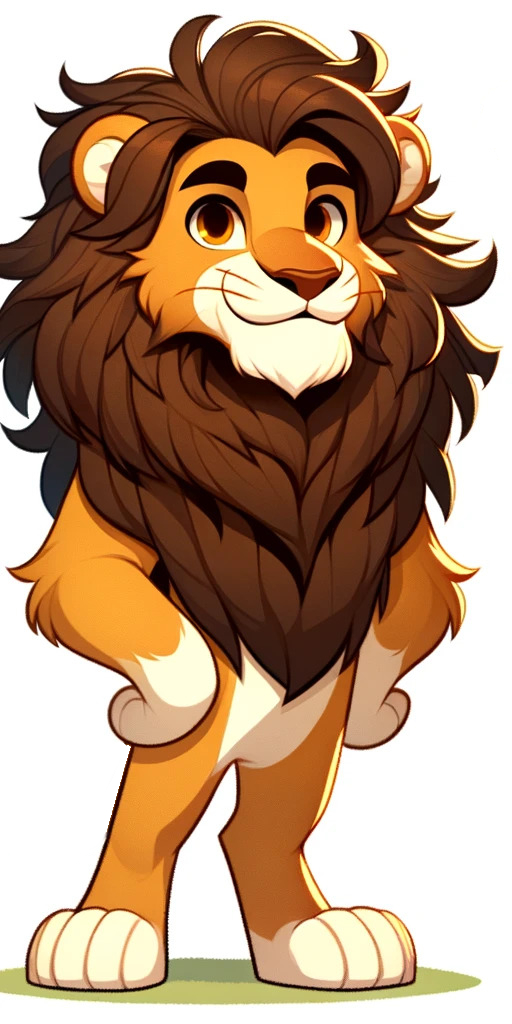}
    \subcaption{}
    \label{fig:subfigB}
  \end{minipage}%
  \begin{minipage}{0.16\columnwidth}
    \centering
    \includegraphics[width=0.8\linewidth]{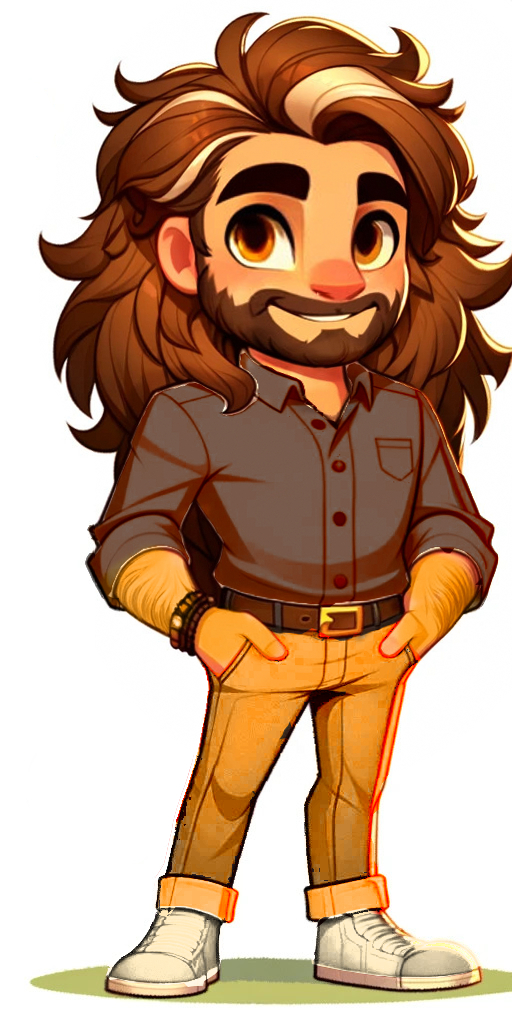}
    \subcaption{}
    \label{fig:subfigC}
  \end{minipage}%
  \begin{minipage}{0.16\columnwidth}
    \centering
    \includegraphics[width=0.8\linewidth]{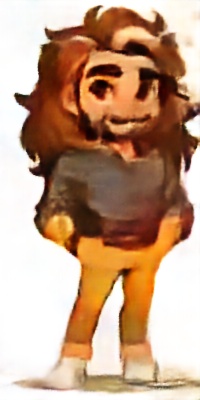}
    \subcaption{Cls: Human}
    \label{fig:subfigD}
  \end{minipage}
  \caption{We take a noisy image (a), which can be interpreted as an animal (b) or human (c). We denoise and classify the image (a), aiming to improve human perception resulting in (d). Note, in a real application (b) and (c) would not be available, which increases the difficulty of interpreting the noisy image. {\scriptsize Artist: DALL-E-2 \cite{DallE2} }}
  \label{fig:main}
\end{figure}


A model combining both denoising and classification is studied in \cite{joint_denoising_classification}, focusing on denoising performance. In addition, we optimize for efficiency, which is required for edge devices, and classification. Our efficiency definition is based on two elements: (i) latency reduction while (ii) retaining denoising performance and classification accuracy. These elements could be optimized using independent classification and denoising models. However, we propose an architecture combining the tasks more efficiently.

First, we employ established model design approaches to enhance independent denoising and classification models, such as model scaling \cite{EfficientNet, UNet_reduction} and 
efficient operators \cite{MobileNetV2}. Although the models are optimized, they still operate separately, resulting in unnecessary overhead. Hence we propose and compare two methods that join both tasks, yielding a novel and efficient architecture. 

Adjusting this architecture for each device and desired latency can be laborious and requires expert knowledge. These issues have recently garnered interest, leading to the emergence of new automated architecture search techniques, which have achieved competitive results in image classification \cite{EfficientNetV2, FB_net}. Moreover, recent Neural Architecture Search (NAS) approaches incorporate latency in their loss function, enabling the design of architectures tailored to specific latency requirements \cite{MobileNetV3, TF_NAS, FB_net}. Combining NAS with the proposed architecture provides a seamless and efficient approach to designing denoising and classification models for diverse use cases. 

We find that our proposed efficiency-focused architecture consistently outperforms our more straightforward one. This is observed for both the manually and NAS designed models. In addition, our NAS models significantly outperform the manually designed ones in denoising and classification performance.

We have the following contributions. (i) We introduce a novel architecture to efficiently combine denoising and classification. The novelty lies in sharing an encoder between the denoiser and the classifier. (ii) We propose modifications to an existing NAS method for classification \cite{TF_NAS} to stabilize its search, improving the performance of the found architectures. (iii) We extend an existing NAS method to search for a model that combines denoising and classification, optimized for a target latency, classification accuracy, and denoising performance.

Since no prior work proposes a joint efficient model for denoising and classification, we study the tasks both separately and joint in Sec. \ref{sec:toy-setup}. The findings are used as expert knowledge to construct the NAS method in Sec. \ref{sec:NAS}.\footnote{Project site: \url{https://thomas-markhorst.github.io}}

%% file: sec/2_related_work.tex
\section{Related work}
\label{sec:related_work}
\textbf{Denoising.}
Image denoising aims to reconstruct a clean image \(x\) from its observed noisy variant \(y\). This relation can be formulated as \(y = x + n\), where we assume \(n\) to be additive white Gaussian noise (AWGN). Neural network-based denoisers offer faster inference and good performance compared to traditional denoising methods like BM3D \cite{BM3D} and WNNM \cite{WNNM}. The interest in deep learning for denoising started with DnCNN \cite{DnCNN}, a simple Convolutional Neural Network (CNN). Encoder-decoder architectures became popular due to their efficient hierarchical feature extraction. Specifically, UNet \cite{UNet} whose skip-connections between the encoder and decoder enhance the denoising process as shown in follow-up methods \cite{RDUNet, MWCNN, DHDN}. The interest in the UNet structure continues with transformer architectures \cite{Uformer, SUNet}. In this paper, our denoisers are based on UNet, ensuring our findings can translate to most related work.

\textbf{Efficient classification.} Optimization for efficiency is generally achieved by either compressing pre-trained networks \cite{compression_survey} or designing small networks directly \cite{MobileNetV2, EfficientNetV2}. We focus on efficient design, for which handcrafted models and neural architecture search (NAS) play essential roles. Studies proposing handcrafted models often introduce efficient operators \cite{MobileNetV2, MobileNet, ShuffleNet} or scaling methods \cite{EfficientNet}. These efficient operators are used in NAS methods \cite{EfficientNetV2, FB_net} aiming for the automated design of efficient neural networks. Such an operator is the inverted residual with a linear bottleneck (MBConv), as introduced in MobileNetV2 \cite{MobileNetV2}. 
In our models, we study scaling methods and MBConv's efficiency characteristic.

\textbf{Neural Architecture Search.} The use of reinforcement learning (RL) for neural architecture search introduced efficient architectures with competitive classification performance \cite{EfficientNetV2, MobileNetV3, MNASNet, ENAS}. However, their discrete search space is computationally expensive.
Differentiable NAS (DNAS) methods \cite{DARTS, FB_net, ProxylessNAS} significantly reduce this cost by relaxing the search space to be continuous using learnable vectors \(\alpha\) for selecting candidate operations, which allows for gradient-based optimization. The popularity of DNAS started with DARTS \cite{DARTS}, which searches a cell structure. 
Due to the complex design and repetitiveness throughout the network of the cell structure, follow-up works \cite{FB_net, TF_NAS} search operators for every layer instead of constructing repeating cells. 

Pitfalls of DNAS are the collapse of search into some fixed operations and a performance drop when converting from the continuous search network to the discretized inference network \cite{FairDARTS, SNAS, BDARTS}. TF-NAS \cite{TF_NAS} addresses these issues with an adaptation in the search algorithm, which lets the search model mimic the discrete behavior of the inference model. 
In addition, TF-NAS searches an architecture with a target latency by adding a latency loss to the search optimization. Because of these properties, we use TF-NAS as a baseline for our NAS study.

Existing NAS methods for denoising are either not reproducible \cite{SuperkernelNAS_denoising}, have a cell-based search space \cite{HiNAS_denoising}, or do not have an encoder-decoder \cite{DPNAS_denoising} architecture. Instead, we use a layer-based search space and encoder-decoder structure.

\textbf{Joint classification and denoising.} In \cite{joint_relation}, the positive influence of denoising methods on classification performance is discussed. Moreover, \cite{joint_denoising_classification} proposed a joint model where a VGG classifier \cite{VGG} is attached to a denoiser similar to UNet. This method \cite{joint_denoising_classification} reports a qualitative improvement of the denoised images when adding the classification loss to the denoiser's optimization, whereas \cite{joint_relation} reports a quantitative improvement. Although these models denoise and classify well, they are not optimized for efficiency. In this paper, we design a joint image denoising and classification method for edge devices.

%% file: sec/3_toy_setup.tex
\section{Exploiting Expert Knowledge}
\label{sec:toy-setup}
\begin{figure}[t]
     \centering
     \centering    \includegraphics[width=0.7\columnwidth]{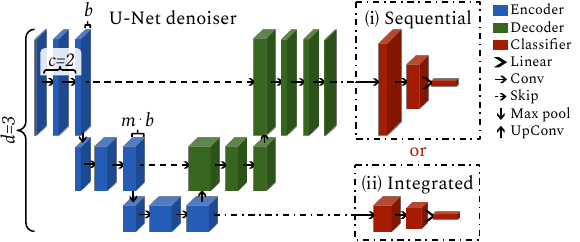}
    \caption{A UNet, with hyperparameters base feature map width (\textit{b}), depth (\textit{d}), channel multiplier (\textit{m}) and convolutions per layer (\textit{c}). For the joint model either attach the classifier to form (i) the Sequential model or (ii) the Integrated model.}
    \label{fig:integrated}
\end{figure}

We start in a controlled setting with separate baseline models for classification and denoising. Additionally, methods to increase their respective efficiency are studied, resulting in a reduced version of the baseline denoiser and classifier. Both the construction and efficiency improvement of the models are described in Suppl. \ref{app:classification}, where a UNet (Fig. \ref{fig:integrated}) and simple 2-block CNN (Fig. \ref{fig:integrated}{\color{black}.i} and \ref{fig:integrated}{\color{black}.ii}) are used as baseline denoiser and classifier respectively. This section describes how the different sizes of the classifiers and denoisers are used to study joining methods and their efficiency. 

\textbf{Dataset \& settings.} For the experiments in this section, we generate a controlled synthetic data set to study the behavior of the classifier and denoiser when applying model scaling, replacing the convolutional operations, and combining both models. The dataset consists of 30k images, each with a random constant background in a gray tint [0.1 - 0.3] with two randomly placed non-overlapping MNIST \cite{MNIST} digits. We use two digits to increase the complexity of the denoising task. For experiments including classification, the two digits are extracted from the image using ground truth locations. These extracted digits are separately used as input for the classifier. In the experiments where noise is required, for either denoising or noisy classification, synthetic Gaussian noise is added. This noise is zero mean, and the intensity of the noise is controlled using the standard deviation (\(\sigma\)) of the distribution. Fig. \ref{fig:sub1} shows a sample, and Fig. \ref{fig:sub3} its noisy variant. To test the model behavior on an extensive noise range, every model is trained and tested on eleven \(\sigma\) values evenly spaced on the interval \([0,1]\) (Tabs. \ref{tab:joint}, \ref{tab:classification-appendix} and \ref{tab:denoising}). The models are trained using Adam optimizer with 1E-3 learning rate (LR), plateau LR scheduler, and 100 epochs.


Since the experiments with the controlled data set are not targeted at a specific device, the metric defining efficiency should not depend on a device. Such a metric is computational power, most commonly defined as Floating Point Operations (FLOPs), which we use as the primary metric. Despite being device dependent, we assess latency as a secondary metric. The latency is measured with a batch size of 32, 100 warm-up inference passes and averaged over 1000 inference passes. Classification performance is quantified using accuracy, while for denoising performance the Peak Signal-to-Noise Ratio (PSNR) and Structural Similarity Index (SSIM) metrics \cite{SSIM_PSNR} are used. 
Higher is better for all our metrics.

\subsection{Joint model: DC-Net}
\label{subsec:dc-net}
\textbf{Experimental setup.} We construct a baseline and reduced joint model, Denoising-Classifying Network (DC-Net). Both the baseline and reduced DC-Net use the same classifier (MB2.5-M Suppl. \ref{app:classification}). Whereas UNet-S and UNet are used for the reduced and baseline DC-Net respectively.

For joining the denoiser and classifier, we propose two models: (i) a Sequential model where the classifier is attached after the denoiser (Fig. \ref{fig:integrated}{\color{black}.i}), and (ii) an Integrated model, the classifier is attached to the UNet encoder (Fig. \ref{fig:integrated}{\color{black}.ii}). For the Integrated model, classification and denoising branches share the encoder.

The benefits of the Integrated model could come in threefold. First, using a shared encoder removes the need for a second large classifier, as in the Sequential method. Second, the decoder and classifier branches could run in parallel compared to running sequentially, which can result in lower latency. Thirdly, the decoder is only optimized for denoising, since the optimization of the classifier does not influence it anymore. It can result in better image quality.

The models are trained using a weighted combination of the Cross-Entropy and Charbonnier loss \cite{AIMWinnerCharbonnier, CharbonnierBaron} (Eq. \ref{eq:DC-Net}). We report the metrics averaged over all 11 noise levels, \(\sigma\) in [0, 1]. 

\begin{equation}
    \mathcal{L} = 0.1 \cdot \mathcal{L}_\text{CE} + 0.9 \cdot \mathcal{L}_\text{Char}
    \label{eq:DC-Net}
\end{equation}

\begin{figure}[t]
  \centering
  \begin{subfigure}{0.13\columnwidth}
    \includegraphics[width=\linewidth]{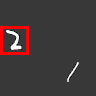}
    \caption{GT}
    \label{fig:sub1}
  \end{subfigure}
  \begin{subfigure}{0.13\columnwidth}
    \includegraphics[width=\linewidth]{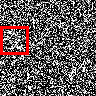}
    \caption{\(\sigma\)=0.8}
    \label{fig:sub3}
  \end{subfigure}
  \begin{subfigure}{0.13\columnwidth}
    \includegraphics[width=\linewidth]{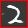}
    \caption{\textbf{Int. S}}
    \label{fig:sub8}
  \end{subfigure}
  \begin{subfigure}{0.13\columnwidth}
    \includegraphics[width=\linewidth]{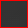}
    \caption{Seq. S}
    \label{fig:sub9}
  \end{subfigure}
  \begin{subfigure}{0.13\columnwidth}
    \includegraphics[width=\linewidth]{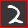}
    \caption{\textbf{Int. L}}
    \label{fig:sub10}
  \end{subfigure}
  \begin{subfigure}{0.13\columnwidth}
    \includegraphics[width=\linewidth]{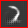}
    \caption{Seq. L}
    \label{fig:sub11}
  \end{subfigure}
  
  \caption{Ground-truth sample (a), which is the target for the denoiser when given noisy image (b). S stands for the reduced model and L for the baseline. (c-f) are the cropped denoised outputs for input (b) and the red squares indicate the zoomed-in regions. For higher noise levels, the denoising performance of the Sequential model is worse than the Integrated model.
  }
  \label{fig:joint}
\end{figure}

\textbf{Exp 1. Integrated vs. Sequential.} Which joining method performs better for the baseline, and does the same conclusion hold when reducing its size? We compare the \textit{Sequential} and \textit{Integrated} models. In Tab. \ref{tab:joint}, we see that for both the baseline and reduced DC-Net models, the Integrated version performs significantly better at denoising, while the Sequential version performs better at classification. The difference in denoising performance is visualized in Fig. \ref{fig:joint}. We see that both the reduced (\ref{fig:joint}\textcolor{black}{c}) and baseline (\ref{fig:joint}\textcolor{black}{e}) Integrated models reconstruct the digit clearly. Whereas both sizes of the Sequential model (\ref{fig:joint}\textcolor{black}{d} and \textcolor{black}{f}) fail to reconstruct the digit. 

\begin{table}[b]
\centering
\caption{Comparison of the reduced and baseline joint models. Both the Integrated and Sequential methods trained on the synthetic noise dataset. The integrated model performs significantly better in denoising and slightly worse in classification. The integrated model also scales down better.}
\resizebox{0.80\columnwidth}{!}{%
\begin{tabular}{@{}ccccccc@{}}
\toprule
DC-Net & Type & FLOPs (M) \(\downarrow\) & Lat. (ms) \(\downarrow\) & PSNR \(\uparrow\) & SSIM \(\uparrow\) & Acc. (\%) \(\uparrow\)\\ \midrule
\multirow{2}{*}{Baseline} & \textbf{Integrated} & 1301.8 & \textbf{7.14} & \textbf{32.8} & \textbf{0.97} & 88.1 \\
 & Sequential & 1302.1 & 7.55 & 27.1 & 0.95 & \textbf{89.6} \\ 
\cmidrule(r){1-2}
\multirow{2}{*}{Reduced} & \textbf{Integrated} & 51.2 & \textbf{2.41} & \textbf{29.9} & \textbf{0.97} & 86.2 \\
 & Sequential & 51.5 & 2.83 & 25.2 & 0.92 & \textbf{87.6} \\ 
\bottomrule
\end{tabular}%
}
\label{tab:joint}
\end{table}

\textbf{Conclusion.}
The integrated model has a slightly lower classification accuracy compared to the Sequential model, yet it has superior performance in terms of image quality. When aiming for improved human perception, it is still required for the human to see the content of the image. Therefore, the Integrated model is more suitable for joint denoising and classification and is called DC-Net.

%% file: sec/4_nas.tex
\section{Neural Architecture Search}
\label{sec:NAS}
We follow similar experimentation strategies as in the previous section. TF-NAS\cite{TF_NAS} is used to construct a classifier, which we use as a basis for our denoiser and joint model. All our proposed models in this Section contain searchable blocks, the models and which parts are searchable are defined in Figure \ref{fig:DC-NAS}.

\textbf{Dataset \& settings.}
The following experiments are conducted on Imagenet \cite{Imagenet}, randomly cropped
    to 224x224 pixels. To reduce search and training time, 100 classes (Imagenet 100) from the original 1000 classes were chosen, as in \cite{TF_NAS}. In the experiments requiring noise, Gaussian noise is sampled uniformly with a continuous range of \(\sigma\) in \([0,1]\) (Tabs. \ref{tab:NAS-denoising}, \ref{tab:main-results}, \ref{tab:ablation1}, \ref{tab:ablation2} and \ref{tab:ablation3}).

The models are searched using SGD with momentum, 2E-2 LR with 90 epochs. Afterward, the found architecture is trained from scratch with 2E-1 LR for 250 epochs. All other settings are similar to \cite{TF_NAS}. The loss function depends on the task of the experiment, Cross-Entropy with label smoothing for classification ($\mathcal{L}_\text{CE}$), combined Charbonnier and SSIM losses for denoising ($\mathcal{L}_\text{Den}$), and a weighted combination for the joint model ($\mathcal{L}_\text{Both}$), see Eq. \ref{eq:DC-NAS_Den}, \ref{eq:DC-NAS_Both}.

\begin{align}
    \mathcal{L}_\text{Den} = 0.8 \cdot \mathcal{L}_\text{Char} + 0.2 \cdot \mathcal{L}_\text{SSIM} \label{eq:DC-NAS_Den}\\
    \mathcal{L}_\text{Both} = 0.1 \cdot \mathcal{L}_\text{CE} + 0.9 \cdot \mathcal{L}_\text{Den}\label{eq:DC-NAS_Both}
\end{align}

Since our NAS method uses a latency look-up table constructed for our device, these experiments target a specific device, GeForce RTX 3090 GPU. Therefore latency is suitable for defining efficiency in the NAS experiments.

\begin{figure*}[t]
   \begin{minipage}{0.363\textwidth}
        \centering
        \includegraphics[width=\textwidth]{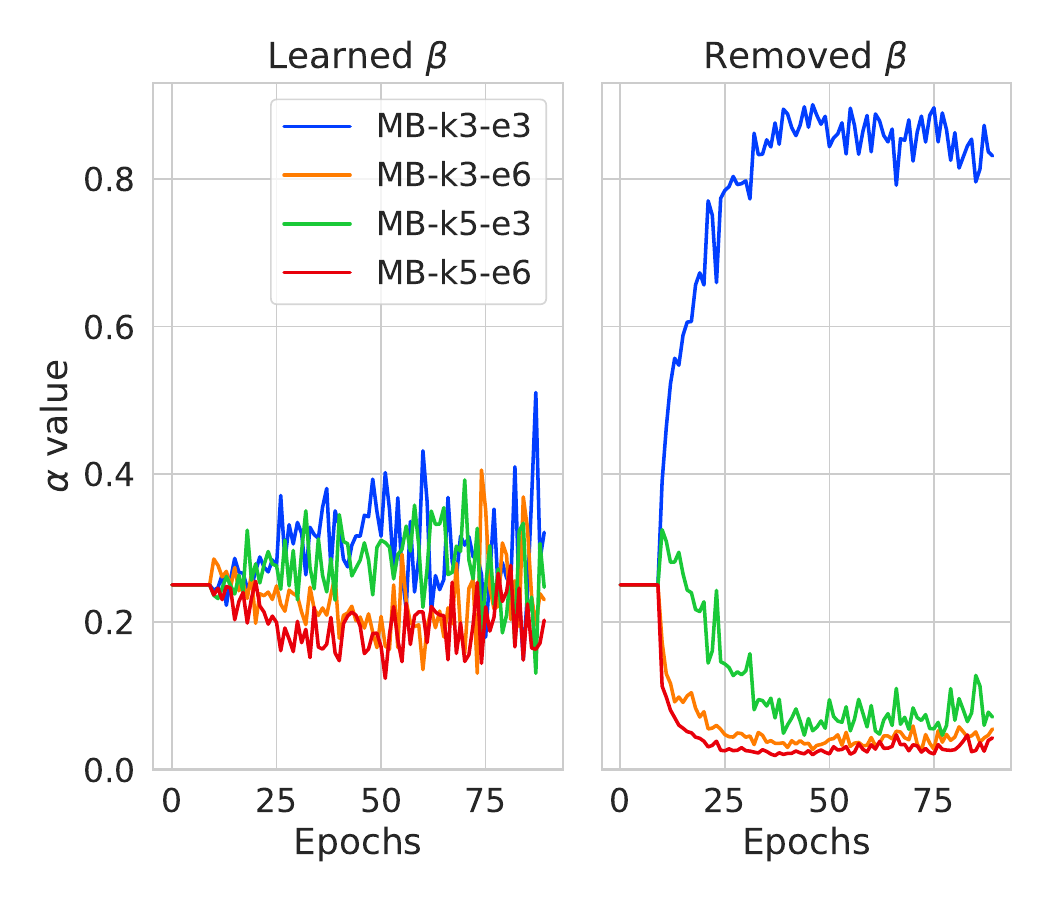}
        \captionsetup{width=.95\textwidth}
        \vspace{-0.8cm}
        \caption{Stage-5:block-4's $\alpha$ values for Removed and Learned $\beta$. Search is more stable when $\beta$ is removed.\\}
        \label{fig:NAS-beta}
   \end{minipage}\hfill
   \begin{minipage}{0.314\textwidth}
        \centering
        \includegraphics[width=\textwidth]{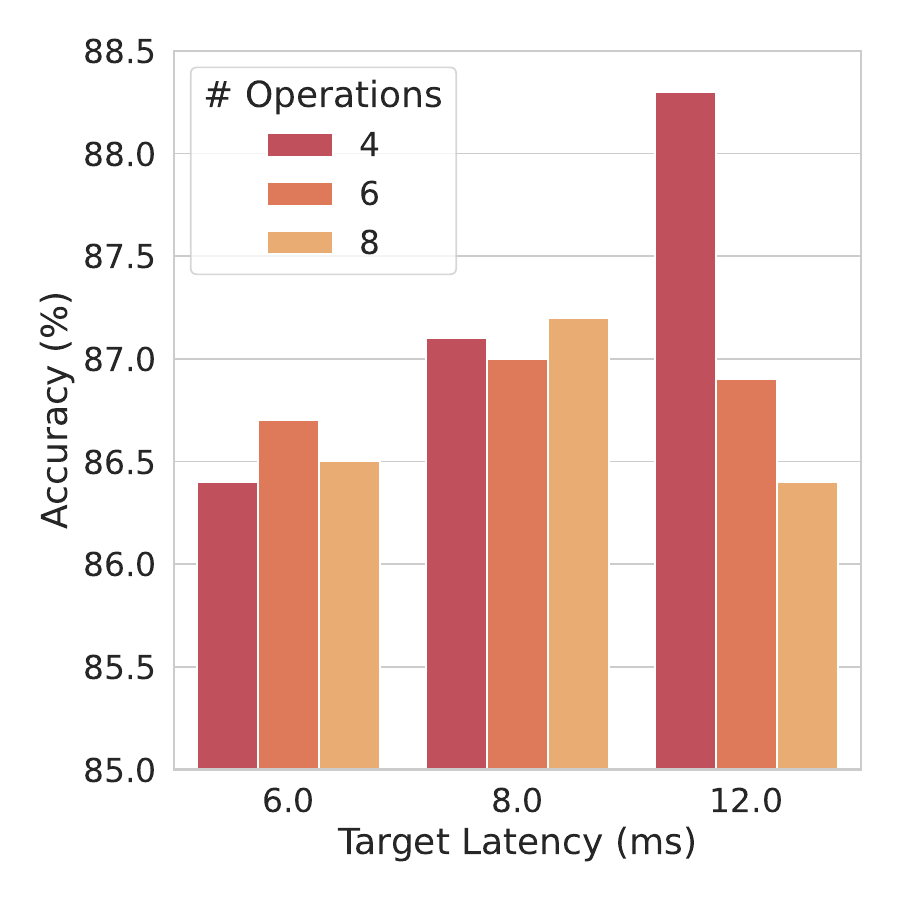}
        \captionsetup{width=.95\textwidth}
        \vspace{-0.8cm}
        \caption{Acc for different search spaces, showed for different target latencies. Using fewer operations is the most robust.}
        \label{fig:NAS-number-blocks}
   \end{minipage} \hfill
   \begin{minipage}{0.314\textwidth}
        \centering
        \includegraphics[width=\textwidth]{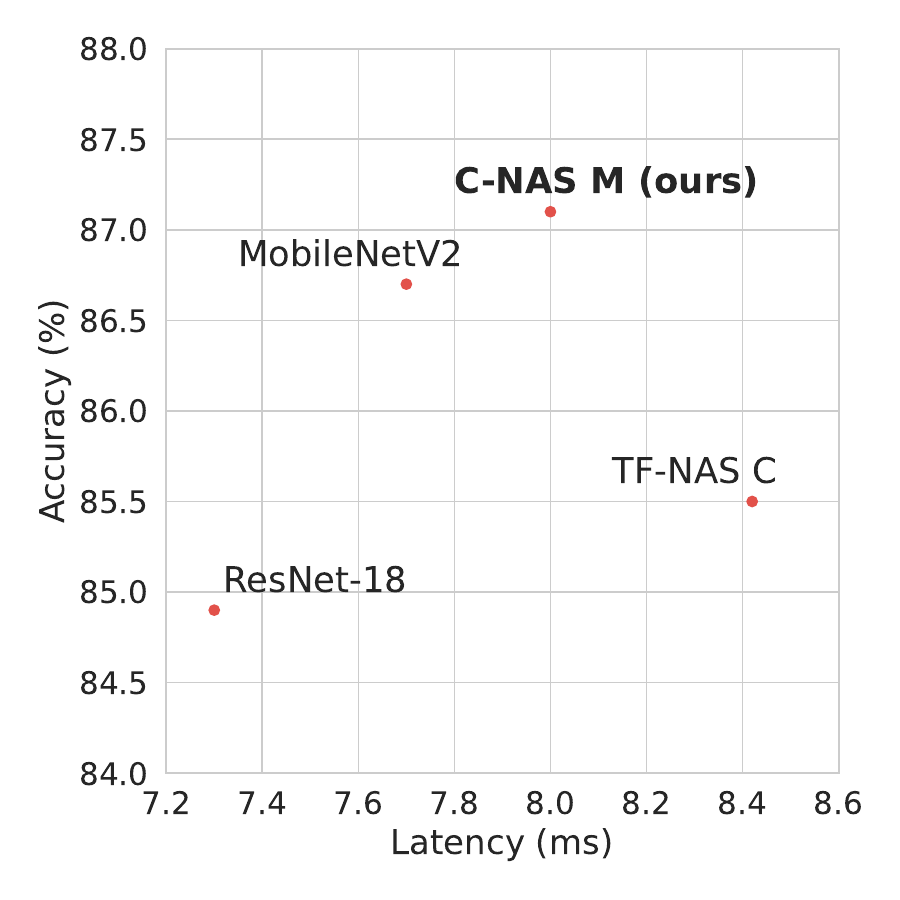}
        \captionsetup{width=.95\textwidth}
        \vspace{-0.8cm}
        \caption{Comparing classifiers with similar latency on Imagenet100. Our model outperforms other methods. }
        \label{fig:NAS-classification}
   \end{minipage}
\end{figure*}

\subsection{Classification: C-NAS}
\label{subsec:NAS-classification}
\textbf{Experimental Setup.}
Since TF-NAS \cite{TF_NAS} learns \(\beta\)'s to control the number of convolutional operators per stage, \(\beta\)'s can reduce the model size. However, in the models proposed by \cite{TF_NAS}, only 2 out of 24 stages are reduced by \(\beta\). So the \(\beta\)'s have little effect on the found architectures, yet they make the search space more complex. Therefore, we propose a version of TF-NAS where the \(\beta\)'s are removed so that all convolutional blocks are used. 

The candidate operations in the search space of TF-NAS are MBConvs with 8 different configurations, see Suppl. \ref{app:search-space}. The configurations differ in kernel size, expansion rate, and in- or excluding a squeeze- and excitation layer (SE) \cite{SE-layer}. 

The classification experiments are performed using data without noise, as the aim is to examine the NAS method, which is designed for clean images. We investigate key components of TF-NAS and try to improve its stability and classification performance.

\textbf{Exp 1. Learned vs. Removed \(\beta\).} We conduct an experiment to study the effect of removing \(\beta\) on the search quality. The SE-layer is excluded from the candidate blocks, halving the search space to ensure the number of candidate operations does not cause search instability. We set a low target latency of 6 ms, as learning \(\beta\) should have a positive effect on small networks. For both the learned and removed settings, we run two searches, search 1 and 2.

Fig. \ref{fig:NAS-beta} shows that when \(\beta\) is learned, the \(\alpha\)'s selecting a candidate operation oscillate and therefore do not decide on an architecture. Whereas with Removed \(\beta\), the search is stable. This stability is reflected in the performance, as the average accuracy of the Removed \(\beta\) models is 86.3\%, compared to 84.2\% for Learned \(\beta\). The separate results for each model are shown in Suppl. \ref{app:nas-beta}.

\textbf{Exp 2. Number of operators in search space.}
Does reducing the number of operators during search positively influence the performance of the found models? We test this by comparing the performance of architectures searched with three different search space sizes, \{4, 6, or 8\} operations, defined in Suppl. \ref{app:search-space}. For each of these search spaces, three different latency targets are used: \{6, 8, and 12\} ms. 

In Fig. \ref{fig:NAS-number-blocks}, we see that for lower target latencies, 6 and 8 ms, using fewer operations in the search space does not alter performance significantly. When targeting 12 ms latency, reducing the number of operations in the search space does show a significant improvement.
Additionally, we find that when using the larger search spaces, the operators from the small search space are still preferred for lower latencies.

\textbf{Exp 3. Compare with original TF-NAS.} How do architectures found using our proposed changes to TF-NAS perform compared to models with similar latency? We compare our model, C-NAS M, with TF-NAS C, MobileNetV2, and ResNet-18. MobileNetV2 our model have similar latency, architecture, and operator types. ResNet only differs in that it uses the Conv operator. We include these standard baseline architectures to indicate where C-NAS, see Fig. \ref{fig:DC-NAS}{\color{black}.i}, stands on Imagenet100.

Fig. \ref{fig:NAS-classification} shows that the model found using our method has lower latency yet higher accuracy than TF-NAS C as proposed in \cite{TF_NAS}. The model is searched with target latency 8.0. We observe that our search method is able to find a model that matches its target latency. Although ResNet-18 and MobileNetV2 run faster than our model, our classification accuracy is superior, especially when compared to ResNet-18, which only uses Convs.

\textbf{Conclusion.} By Removing \(\beta\) and reducing the number of operators used in the search, the search stability increases, and we find architectures that have better accuracy. An architecture found using our changes classifies better than a TF-NAS architecture with similar latency.

The comparison between our model and ResNet-18 shows that our search space is able to compete with widely accepted Conv-based classifiers. Moreover, our model performs on par with MobileNetV2, a manually designed classifier using MBConvs.

\subsection{Denoising: D-NAS}
\textbf{Experimental setup.}
To construct a denoiser, D-NAS (Fig. \ref{fig:DC-NAS}{\color{black}.ii}), we use the first six stages of a found C-NAS classifier, which has four levels of resolution. Afterwards, we attach a UNet style decoder by using both a transposed convolution and two normal convolutions for each decoder level. Like UNet, we also add skip connections between the encoder and decoder layers. The decoder is not searched.

\textbf{Exp 1. D-NAS vs UNet denoiser.}
Does our denoiser D-NAS perform similarly to the UNet denoisers? For this experiment, we use UNet-S (Sec. A.2) \{\textit{d} = 4, \textit{b} = 8, \textit{c} = 2, \textit{m} = 1.5\}, with a latency of 9.2 ms and the larger UNet-M,  \{\textit{d} = 4, \textit{b} = 16, \textit{c} = 2, \textit{m} = 2\} with a latency of 16.9 ms.  We compare them with our D-NAS M, with similar latency.

\begin{table}[]
\caption{Comparison of D-NAS and UNet variants for denoising. D-NAS outperforms slightly faster UNet-S, but UNet-M denoises best at the cost of 45\% higher latency.} 
\centering
\resizebox{0.55\columnwidth}{!}{%
\begin{tabular}{@{}lccccccc@{}}
\toprule
\multirow{2}{*}{Model} & \multicolumn{3}{c}{UNet params:} & \multirow{2}{*}{Lat. (ms) \(\downarrow\)} & \multirow{2}{*}{PSNR \(\uparrow\)} & \multirow{2}{*}{SSIM \(\uparrow\)} \\
 & d & b & m &  &  &  &  \\ \midrule
UNet-S & 4 & 8 & 1.5 & 9.2 & 25.0 & 0.69 \\
UNet-M & 4 & 16 & 2 & 16.9 & 25.9 & 0.72 \\ 
D-NAS M & - & - & - & 11.6 & 25.6 & 0.71 \\
\bottomrule
\end{tabular}%
}
\label{tab:NAS-denoising}
\end{table}

Tab. \ref{tab:NAS-denoising} shows that D-NAS M outperforms UNet-S by 0.6 dB PSNR and 2\% SSIM, at the cost of 2.4 ms latency. However, the 7.7 ms slower UNet variant, UNet-M, denoises better than our proposed model, by 0.3 dB and 1\% SSIM.

\textbf{Conclusion.} D-NAS performs similarly to our baseline UNets. Therefore D-NAS is a suitable denoising architecture and it can form the backbone of our Integrated model.

\begin{table}
\caption{Comparison of DC-NAS models searched for three different latencies, with their corresponding C-NAS model, classifier baseline, and denoiser baseline. Our Integrated models perform similar or better than their corresponding baselines, with the advantage of having a joint denoising and classification network. *See Suppl. \ref{subsubsec:eff-den}.}
\centering
\resizebox{0.8\textwidth}{!}{%
\begin{tabular}{@{}llcccc@{}}
\toprule
\multicolumn{1}{c}{\multirow{2}{*}{Model}} &
  \multicolumn{1}{c}{\multirow{2}{*}{Type}} &
  \multirow{2}{*}{Lat. (ms) \(\downarrow\)} &
  Classification &
  \multicolumn{2}{c}{Denoising} \\ \cmidrule(l){4-6} 
\multicolumn{1}{c}{} &
  \multicolumn{1}{c}{} &
   &
  Acc. (\%) \(\uparrow\) &
  PSNR \(\uparrow\) &
  SSIM \(\uparrow\) \\ \midrule
MobileNetV3 \cite{MobileNetV3}                     & Classifier & 4.9  & 70.4 & -             & -    \\
C-NAS S (\textit{ours})                                      & Classifier & 5.9  & 73.5 & -             & -    \\
UNet-S* \cite{UNet} & Denoiser   & 9.2  & -    & 25.0          & 0.69 \\
DC-Net S (\textit{ours})                                     & Integrated & 10.0 & 61.9 & 24.5          & 0.68 \\ 
DC-NAS S (\textit{ours})                                     & Integrated & 10.3 & \textbf{74.3} & \textbf{25.4}          & \textbf{0.70} \\ \midrule
EfficientNetV2-b0 \cite{EfficientNetV2}            & Classifier & 9.0  & 75.4 & -             & -    \\
C-NAS M (\textit{ours})                                      & Classifier & 7.9  & 75.5 & -             & -    \\
LPIENet 0.25x \cite{LPIENet}                        & Denoiser   & 12.7 & -    & 24.1          & 0.65 \\
DC-NAS M (\textit{ours})                                     & Integrated & 13.7 & \textbf{76.0} & \textbf{25.4}          & \textbf{0.70} \\ \midrule
EfficientNetV2-b1 \cite{EfficientNetV2}            & Classifier & 11.8 & \textbf{76.7} & -             & -    \\
C-NAS L (\textit{ours})                                      & Classifier & 12.0 &   76.0   & -             & -    \\
UNet-M* \cite{UNet}  & Denoiser   & 16.9 & -    & \textbf{25.9 }         & \textbf{0.72} \\
LPIENet 0.5x \cite{LPIENet}                         & Denoiser   & 19.8 & -    &       24.7        &   0.68   \\
DC-NAS L (\textit{ours}) & Integrated & 17.9 & 76.4 & 25.2 & 0.70 \\ \bottomrule
\end{tabular}%
}
\label{tab:main-results}
\end{table}

\subsection{Joint Model: DC-NAS}
\label{subsec:nas-joint}
\begin{figure}[t]
    \centering
    \includegraphics[width=0.55\columnwidth]{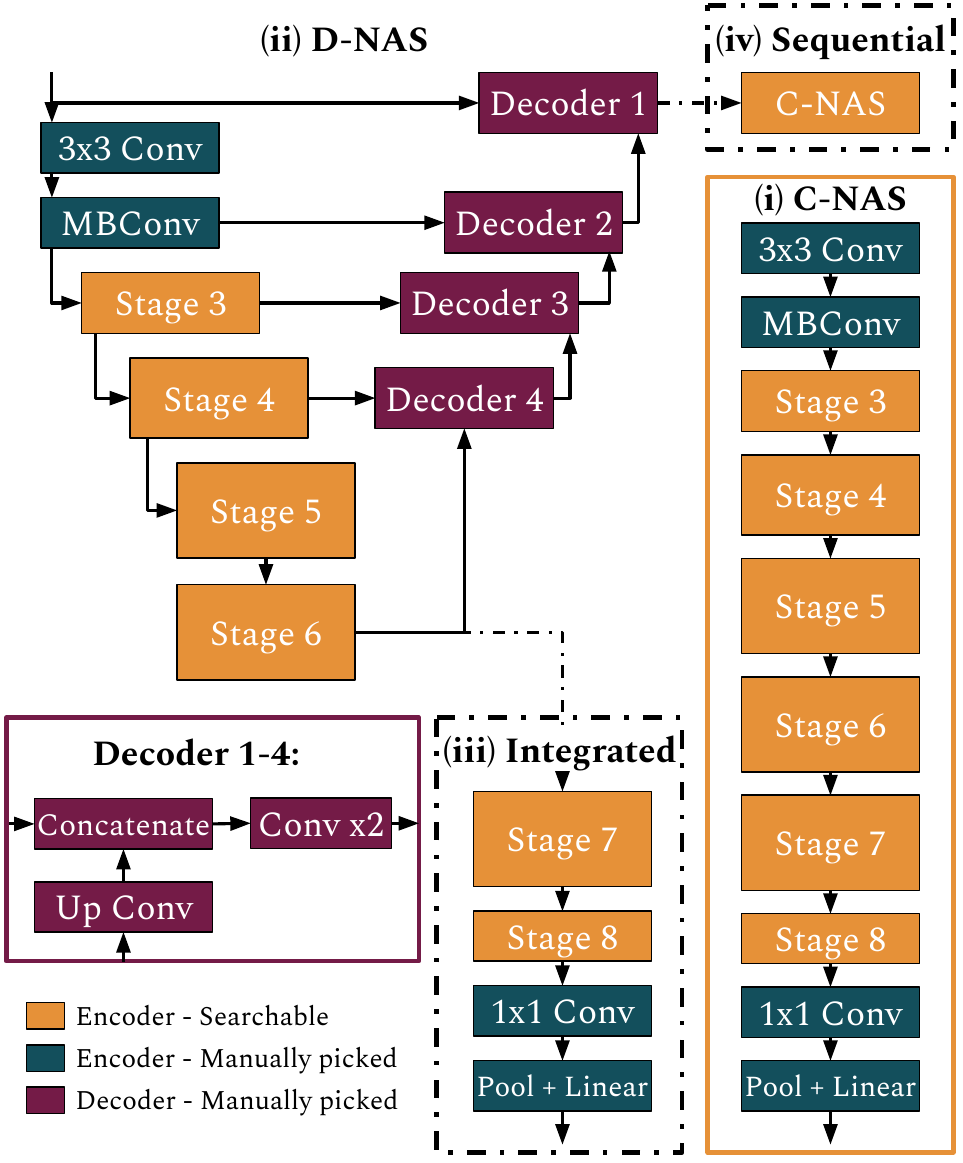}
    \caption{C-NAS and D-NAS architecture. Connecting block (iii) for DC-NAS and block (iv) for DC-NAS$_\text{seq}$. During search models with various latencies can be obtained. Only the orange stages are searchable in the encoder and classifier. }
    \label{fig:DC-NAS}
\end{figure}

\textbf{Experimental setup.}
To construct the joint model, we use the Integrated setup. The Integrated model, DC-NAS, is constructed similarly to D-NAS. We connect the decoder after the first six stages of C-NAS  (Fig. \ref{fig:DC-NAS}, but still use the remaining C-NAS stages(Fig. \ref{fig:DC-NAS}{\color{black}.iii})) as a classification branch. The design choices for DC-NAS are discussed in the ablations study (Sec. \ref{subsec:ablation}).

Using our search method, we search for DC-NAS models of three different sizes \{S, M, L\}. Apart from our manually designed Integrated model, we compare our searched models with separate state-of-the-art classifiers and denoisers, as there are no existing models that jointly optimize denoising, classification, and efficiency. For each DC-NAS model, we also separately train the classifier (C-NAS) to evaluate the influence of joint denoising. The classifier and denoiser baselines are chosen to have similar latency as the C-NAS or D-NAS model on which the corresponding DC-NAS is based.

\textbf{Results.} We discuss the results in Tab. \ref{tab:main-results} in three separate sections for the different target latencies. Our smallest Integrated model, DC-NAS S, outperforms both of its classifier baselines MobileNetV3 and C-NAS S. Note, that the latter shows that the classifier, C-NAS S, performs better when integrated into DC-NAS S. Moreover, our Integrated model denoises better than its baseline UNet-S (Suppl. \ref{app:classification}). DC-NAS S also significantly outperforms our manually designed DC-Net S (Reduced), which is the only Integrated baseline. We display the denoising results of DC-Net, UNet-S, and DC-NAS S in Fig. \ref{fig:visual-comparison}. We observe better denoising on smooth areas, sharper edges, and more realistic color reconstruction for DC-NAS S.

The results of DC-NAS M follow a similar pattern, where our DC-NAS outperforms its baseline denoiser and classifier, using C-NAS M in the Integrated model boosts accuracy by 0.5\%. When comparing DC-NAS M to DC-NAS S, the classification performance improves by 1.7\%, yet the denoising performance plateaus. LPIENet denoises the worst. Comparing DC-NAS M's denoising performance to 3.2 ms slower UNet-M we observe slightly worse denoising performance. However, our Integrated model denoises and classifies with lower latency.

For DC-NAS L, we observe that both the classification and denoising baseline slightly outperform our Integrated model. EfficientNetV2-b1 has 0.3\% higher classification accuracy than DC-NAS L, and UNet-M improves with 0.7 dB PSNR and 2\% SSIM. However, our Integrated model performs both denoising and classification at a similar latency as UNet-M, which only denoises. When comparing DC-NAS L with DC-NAS M, we again note an improvement in classification performance. However, the PSNR score drops by 0.2 dB while the more important SSIM score remains at 0.70.


\textbf{Conclusion.} Our results demonstrate that the proposed DC-NAS models perform similar or better than their denoising and classification baselines for their target latency. In addition, the searched model performs better than our manually designed joint denoiser and classifier.

\subsection{Ablation Study}
\label{subsec:ablation}
\textbf{Exp 1. Encoder search.} C-NAS forms the encoder of DC-NAS and contains the searchable operations within DC-NAS. We test multiple search approaches: (i) using clean images, and (ii) using noisy images. 
For both approaches, we search the encoder using only classification loss $\mathcal{L}_\text{Cls}$. In addition, we also search the DC-NAS encoder on noisy images using the combined denoising and classification loss $\mathcal{L}_\text{Both}$. Therefore it searches for the optimal encoder for both tasks within the Integrated model DC-NAS. Regardless of the search method, the found models are trained using noisy images and the combined loss.

Tab. \ref{tab:ablation1}, shows that when using $\mathcal{L}_\text{Cls}$ with noisy images during search improves classification accuracy by 0.3\%. Surprisingly, the denoising performance is the same. Using both the denoising and classification objectives during the search reduces the classification accuracy. Caused by the denoising loss complicating the search, without improving denoising performance. Therefore, we search our DC-NAS models by only using $\mathcal{L}_\text{Cls}$ loss.

\textbf{Exp 2. Compare Integrated vs. Sequential.} We compare DC-NAS and DC-NAS$_\text{seq}$ models with similar latency. Where DC-NAS$_\text{seq}$ is our Sequential model, which is constructed by attaching C-NAS to the output of D-NAS, see Fig. \ref{fig:DC-NAS}{\color{black}.iv}. Since the searched classifier is used twice in DC-NAS$_\text{seq}$, the Sequential model has a higher latency than the Integrated variant. To counter this, a smaller C-NAS model is used in both the encoder and classifier of DC-NAS$_\text{seq}$. The classifier, C-NAS, used to construct DC-NAS$_\text{seq}$ L has a latency of 6.7 ms. Whereas in DC-NAS L, the classifier has a latency of 12 ms. Note, that these models were searched by using clean instead of noisy images, as this holds for both models it is still a fair comparison.

We see that both models have similar latency and the same classification accuracy, however, DC-NAS L improves denoising performance with 0.5 dB PSNR and 1\% SSIM (Tab. \ref{tab:ablation2}). This improvement is caused by DC-NAS L's Integrated design as this allows for a bigger encoder without increasing latency.

\begin{table}
\caption{Different search strategies for DC-NAS, using (i) clean or noisy images and (ii) $\mathcal{L}_\text{Cls}$ or $\mathcal{L}_\text{Cls} + \mathcal{L}_\text{Den}$. Searching on \textbf{Noisy} images with only $\mathcal{L}_\text{Cls}$ performs best.}
\centering
\resizebox{0.55\columnwidth}{!}{%
\begin{tabular}{@{}llllll@{}}
\toprule
\multicolumn{2}{c}{Search} &
  \multicolumn{1}{c}{\multirow{2}{*}{Lat. (ms)}} &
  \multicolumn{1}{c}{\multirow{2}{*}{Acc. (\%) \(\uparrow\)}} &
  \multicolumn{1}{c}{\multirow{2}{*}{SSIM \(\uparrow\)}} &
  \multicolumn{1}{c}{\multirow{2}{*}{PSNR \(\uparrow\)}} \\ \cmidrule(r){1-2}
Images & Loss                      & \multicolumn{1}{c}{} & \multicolumn{1}{c}{} & \multicolumn{1}{c}{} & \multicolumn{1}{c}{} \\ \midrule
Clean  & $\mathcal{L}_\text{Cls}$  & 13.9                 & 75.7                 & 25.4                 & 0.70                 \\
Noisy  & $\mathcal{L}_\text{Cls}$  & \textbf{13.7}        & \textbf{76.0}        & 25.4                 & 0.70                 \\
Noisy  & $\mathcal{L}_\text{Both}$ & 13.8                 & 75.5                 & 25.4                 & 0.70                 \\ \bottomrule
\end{tabular}%
}
\label{tab:ablation1}
\end{table}

\begin{table}
\caption{Comparing Sequential and Integrated DC-NAS, classification performance is similar, yet the Integrated model is faster and denoises better.}
\centering
\resizebox{0.55\columnwidth}{!}{%
\begin{tabular}{@{}lcccc@{}}
\toprule
Model                 & Lat. (ms) & Acc. (\%) \(\uparrow\) & PSNR \(\uparrow\) & SSIM \(\uparrow\) \\ \midrule
DC-NAS$_\text{seq}$ L & 18.3      & 76.0                   & 25.0              & 0.69              \\
DC-NAS L              & \textbf{17.9}      & 76.0                   & \textbf{25.5}              & \textbf{0.70}              \\ \bottomrule
\end{tabular}%
}
\label{tab:ablation2}
\end{table}

\newcommand\denoisedwidth{0.185}
\begin{figure*}[h!]
    \centering
    \begin{subfigure}{\denoisedwidth\textwidth}
        \includegraphics[width=\textwidth]{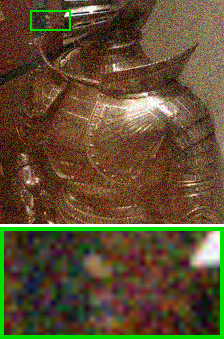}
        \caption{$\sigma$=0.1}
    \end{subfigure}\hfill
    \begin{subfigure}{\denoisedwidth\textwidth}
        \includegraphics[width=\textwidth]{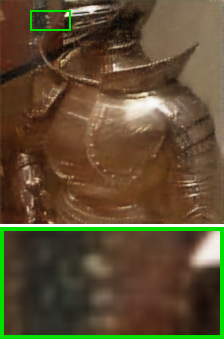}
        \caption{DC-Net}
    \end{subfigure}\hfill
    \begin{subfigure}{\denoisedwidth\textwidth}
        \includegraphics[width=\textwidth]{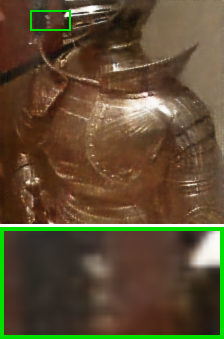}
        \caption{UNet-S}
    \end{subfigure}\hfill
    \begin{subfigure}{\denoisedwidth\textwidth}
        \includegraphics[width=\textwidth]{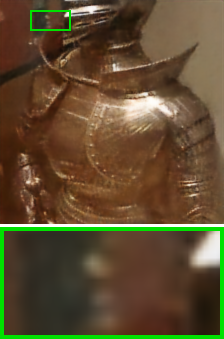}
        \caption{DC-NAS}
    \end{subfigure}\hfill
    \begin{subfigure}{\denoisedwidth\textwidth}
        \includegraphics[width=\textwidth]{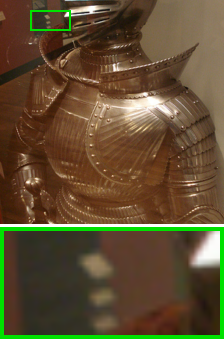}
        \caption{Ground Truth}
    \end{subfigure}

    \vspace{0.2cm}
    \centering
    \begin{subfigure}{\denoisedwidth\textwidth}
        \includegraphics[width=\textwidth]{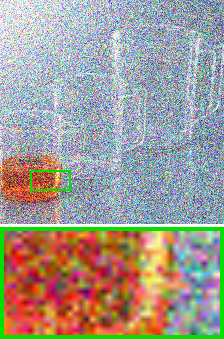}
        \caption{$\sigma$=0.2}
    \end{subfigure}\hfill
    \begin{subfigure}{\denoisedwidth\textwidth}
        \includegraphics[width=\textwidth]{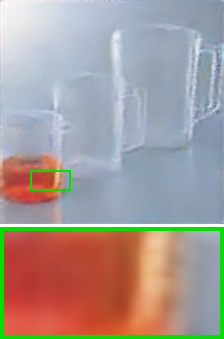}
        \caption{DC-Net}
    \end{subfigure}\hfill
    \begin{subfigure}{\denoisedwidth\textwidth}
        \includegraphics[width=\textwidth]{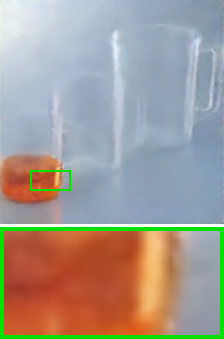}
        \caption{UNet-S}
    \end{subfigure}\hfill
    \begin{subfigure}{\denoisedwidth\textwidth}
        \includegraphics[width=\textwidth]{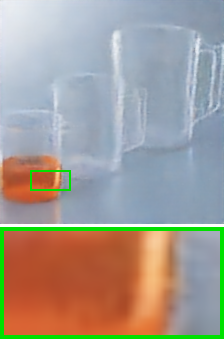}
        \caption{DC-NAS}
    \end{subfigure}\hfill
    \begin{subfigure}{\denoisedwidth\textwidth}
        \includegraphics[width=\textwidth]{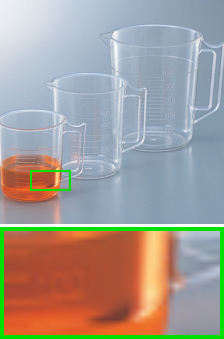}
        \caption{Ground Truth}
    \end{subfigure}

    \caption{Denoising performance of DC-NAS S and its baselines. Left to right: noisy image, the denoiser outputs of size S, and the clean image. Comparing (d) and (b,c), we see better performance in smooth areas and more correct colors in (d). With (i) and (g), we observe a better color reconstruction for (i). Moreover, (i) has less artifacts than (h). Hence, DC-NAS S denoises better than the other denoisers of similar latency.}
    \label{fig:visual-comparison}
    
\end{figure*}

\textbf{Exp 3. Decoder tuning.} The DC-NAS models found in Tab. \ref{tab:main-results} and \ref{tab:ablation2}, have similar denoising performance. These models differ only in the type of MBConvs that are selected during search in the encoder. We test the hypothesis if the denoising performance is influenced by adjusting the operators in the decoder while retaining the latency. DC-NAS M is used as a basis in this experiment. We construct three alternatives. First, the convolutional operators in the decoder are replaced with MBConvs (MB-k3-e3) \cite{MobileNetV2}, which significantly increases the latency of the model. To account for this, we scale down the decoder by (i) using 1 instead of 2 convolutional operations (MBConv) per layer or (ii) using 3 instead of 4 decoder layers. 

In Tab. \ref{tab:ablation3}, we see that using the MBConvs compared to Convs improves the denoising performance. However, at the cost of a 14 ms latency increase, only caused by the MBConv decoder. When reducing the complexity of the MBConv decoder with \textit{1 operator} and \textit{3 layers}, the denoising performance reduces to the original level again, but the latency is still higher than for DC-NAS M which has only standard convolutional layers in the decoder block.

\begin{table}[]
\caption{The influence of altering the Conv operators in the DC-NAS M decoder to MBConv and scaling down the MBConv alternative by reducing the number of operators or decoder layers. Using the standard convolutional layers is more efficient.}
\centering
\resizebox{0.6\columnwidth}{!}{%
\begin{tabular}{@{}lccccc@{}}
\toprule
\multicolumn{2}{c}{Decoder} &
  \multirow{2}{*}{Lat. (ms)} &
  \multirow{2}{*}{Acc. (\%) \(\uparrow\)} &
  \multirow{2}{*}{PSNR \(\uparrow\)} &
  \multirow{2}{*}{SSIM \(\uparrow\)} \\ \cmidrule(r){1-2}
Operator & Scaling    &               &               &      &      \\ \midrule
Conv     & -          & \textbf{13.7}          & 76            & 25.4 & 0.70 \\
MBConv   & -          & 27.7 & 75.5 & 25.8 & 0.71 \\
MBConv   & 1 operator & 16.4          & 75.4          & 25.3 & 0.70 \\
MBConv   & 3 layers   & 22.1          & 75.1          & 25.4 & 0.70 \\ \bottomrule
\end{tabular}%
}
\label{tab:ablation3}
\end{table}

\textbf{Conclusion.} We have seen that the Integrated combining method outperforms its Sequential counterpart in denoising. To construct the integrated model (DC-NAS), we find that searching for a classifier on noisy data, without taking the denoising objective into account results in the best classification performance. Surprisingly, the search method does not influence the denoising performance. Furthermore, manually altering the decoder does not benefit denoising efficiency either. However, the NAS denoising experiments demonstrate that our denoising setup is competitive. Since tuning the decoder operators does not improve performance, our method is focused on searching for only the encoder of the integrated model. The models found by this approach, outperform our manually designed models with similar latency. 

%% file: sec/5_conclusion.tex
\section{Limitations \& Conclusion}
One limitation of our NAS method is its inability to alter the decoder. It is designed this way as manually altering the decoder does not improve efficiency. However, when targeting a significantly different latency, a change in denoising architecture could be required. Therefore, designing model scaling rules for the searched models is of interest, similar to the EfficientNets \cite{EfficientNet, EfficientNetV2}.

Another limitation is the fixation of \(\beta\) in our NAS method. Although this improves the stability of search and network performance, learning \(\beta\) while retaining a stable search would be preferred. This would introduce more possibilities in the search space for optimizing efficiency.

In addition, the latency of Integrated models can be reduced further by running the denoising and classification branches in parallel.

To conclude, we show that using efficient operators and scaling methods proposed in previous work \cite{EfficientNet, UNet_reduction, MobileNetV2} are relevant for denoising and noisy classification. In addition, we present the Integrated model DC-Net to join the two tasks efficiently and show that the Integrated design is more suitable across various latencies than the Sequential variant. To simplify the design process of the joint model when targeting a latency, we present a NAS method. We alter an existing NAS method to improve the stability and performance of the search. This method searches a classifier. Using the searched classifier as a basis, we build the Integrated DC-NAS model. We demonstrate that the proposed model outperforms the manually constructed model. We believe that our study can be a precursor of efficient joint low-level and high-level computer vision tasks.

%% file: sec/X_suppl.tex
\clearpage
\setcounter{page}{1}
\appendix

\section{Efficient Classification \& Denoising: Additional results}
\label{app:classification}
The joint models in Section \ref{sec:toy-setup} are constructed using a separate denoiser and classifier. We describe the baseline models and several methods to construct the reduced versions.

\textbf{Overview of the models used in the main paper.} UNet-S: \{\textit{d} = 4, \textit{b} = 8, \textit{c} = 2, \textit{m} = 1.5\}, which is also called Reduced UNet. UNet-M: \{\textit{d} = 4, \textit{b} = 16, \textit{c} = 2, \textit{m} = 2\}. UNet: \{\textit{d} = 5, \textit{b} = 64, \textit{c} = 2, \textit{m} = 2\}, which is also called Baseline UNet. MB2.5-M: the classifier described in Section \ref{sec:toy-experiments-classification} with an MBConv (expansion rate = 2.5) as second convolutional layer.

\subsection{Efficient Classification}
\label{sec:toy-experiments-classification}
\textbf{Experimental setup.} Our baseline classifier (Conv-L) consists of two convolutional, one global max pooling, and a linear layer. Each convolutional layer also has a group normalization \cite{group_norm}, max pooling, and ReLU activation function.

To construct the reduced version, we use two methods similar to previous works \cite{MobileNetV2, EfficientNet}. In the first method, we replace the second convolutional layer with an MBConv layer. Three expansion rates are used \(\{1, 2.5, 4\}\): (i) rate 1 is the lowest possible value, (ii) rate 4 matches the number of FLOPs of the baseline, and (iii) rate 2.5 is in the middle of those two. The second reduction method is to lower the number of filters in the baseline, also called the model width. Using these techniques, models with three different FLOP sizes are constructed, \{S, M, L\}. We use the following naming scheme, Conv-\(x\) and MB\(e\)-\(x\), where \(x\) represents the FLOP size and \(e\) is the expansion rate of the MBConv.

The models are trained using Cross Entropy loss. We report the accuracy averaged over all 11 noise levels.

\textbf{Exp. 1: Conv vs MBConv comparison.} According to \cite{MobileNetV2}, the MBConv layer should be more efficient than a normal convolutional layer. Therefore, when comparing the two operators in our network, we expect the version with an MBConv layer to need fewer FLOPs for the same accuracy. In Table \ref{tab:classification-appendix}, the MB models with expansion rates 2.5 (MB2.5-M) and 4 (MB4-L) classify better than the Conv-L model with fewer FLOPs. However, with an expansion rate of 1 (MB1-S), the accuracy drops 7\% compared to Conv-L. Therefore, \cite{MobileNetV2}'s theory also holds for the noisy classifier, but only for the higher expansion rates.

\begin{table}
\caption{Classification baseline and reduced models, designed for three different FLOP targets: \{S, M, L\}, to compare scaling methods: expansion rate and model width. Each section of rows is used by the experiments from Sec. \ref{sec:toy-experiments-classification} defined in the \textit{Exp.} column. MB models scale down more efficiently than normal Conv models.}
\centering
\resizebox{0.7\columnwidth}{!}{%
\begin{tabular}{@{}clccccc@{}}
\toprule
Exp. & Model & Size & Exp. rate & FLOPs (K) \(\downarrow\) & Lat. (ms) \(\downarrow\) & Acc (\%) \(\uparrow\)\\ \midrule
\multirow{4}{*}{1-3} & Conv-L & L & - & 447 & 0.336 & 63.2 \\
 & MB1-S & S & 1 & 177 & 0.300 & 56.2 \\
 & MB2.5-M & M & 2.5 & 350 & 0.384 & 64.1 \\
 & MB4-L & L & 4 & 424 & 0.468 & 64.9 \\ \midrule
\multirow{2}{*}{2-3} & MB2.5-S & S & 2.5 & 178 & 0.390 & 58.4 \\
 & MB4-S & S & 4 & 188 & 0.403 & 56.6 \\
 \midrule
\multirow{2}{*}{3} & Conv-S & S & - & 163 & 0.281 & 55.4 \\
 & Conv-M & M & - & 345 & 0.317 & 61.4 \\ \bottomrule
\end{tabular}%
}
\label{tab:classification-appendix}
\end{table}

\textbf{Exp. 2: MBConv width \& expansion rate scaling.} Since MBConv layers can be used to improve efficiency, we question how to further reduce the MB model's FLOP size. We compare two options: (i) reducing the expansion rate and (ii) scaling the width of the network. We take MB4-L as the starting model, as this is our best and largest model.

From the MB models with size S in Table 
\ref{tab:classification-appendix}, MB1-S performs the worst. It only has a reduced expansion rate from 4 to 1. MB4-S, which is obtained by scaling the width of MB-L, increases classification performance by only 0.4\%. However, when slightly reducing MB4-L's width and expansion rate, we derive MB2.5-S, which reaches 58.4\% accuracy, significantly outperforming both other S-sized MB models. So the combination of the two methods is most effective.


\textbf{Exp. 3: Conv width scaling.} In this experiment, we compare the width scaling of the Conv-L model. Table \ref{tab:classification-appendix} shows that all S-sized MB models outperform Conv-S, MB2.5-S even by 3.0\%. MB2.5-M also outperforms Conv-M, by 2.7\%. Therefore, scaling is more efficient for the MB models than the Conv models when optimizing for FLOPs.  




\textbf{Conclusion.}
The MBConv layer can replace the convolutional layers. We find that compared to the Conv models, the MB models also scale down more efficiently by first reducing the expansion rate, possibly followed by a width reduction. Scaling effectively reduces the number of FLOPs.

MB2.5-M has the second-best accuracy with low FLOPs and a latency close to the baseline, Conv-L. Therefore, MB2.5-M is used as the reduced classifier. We also use MB2.5-M as the new baseline classifier as it outperforms the old baseline, Conv-L, in FLOPs and accuracy.

It is important to note that the reduction in FLOPs instantiated by using MBConvs, does not translate to a latency reduction in these experiments. This issue is discussed previously in \cite{EfficientNetV2}. Since the target of these experiments is FLOPs and the latency increase is manageable, we place minimal emphasis on the latency.

\subsection{Efficient Denoising}
\label{subsubsec:eff-den}

\textbf{Experimental setup.} For denoising, the baseline and reduced version are constructed by performing a hyperparameter study on UNet similar to \cite{UNet_reduction}. Figure \ref{fig:integrated} shows the UNet architecture along with its hyperparameters to tune. We explore the parameters one at a time, starting with the number of base features maps \textit{b}, then the UNet depth \textit{d}, the feature map multiplier \textit{m}, and the number of convolutional blocks per layer \textit{c}. In the original UNet: \{\textit{b} = 64, \textit{d} = 5, \textit{m} = 2, \textit{c} = 2\}. Altering these hyperparameters can greatly reduce the model size. Similar to the classification experiments, we also study the ability of the MBConv operator to increase efficiency in the denoiser. The models are trained using Charbonnier loss \cite{charbonnier}. 

\begin{figure}[t]
     \centering
     \includegraphics[width=0.8\columnwidth]{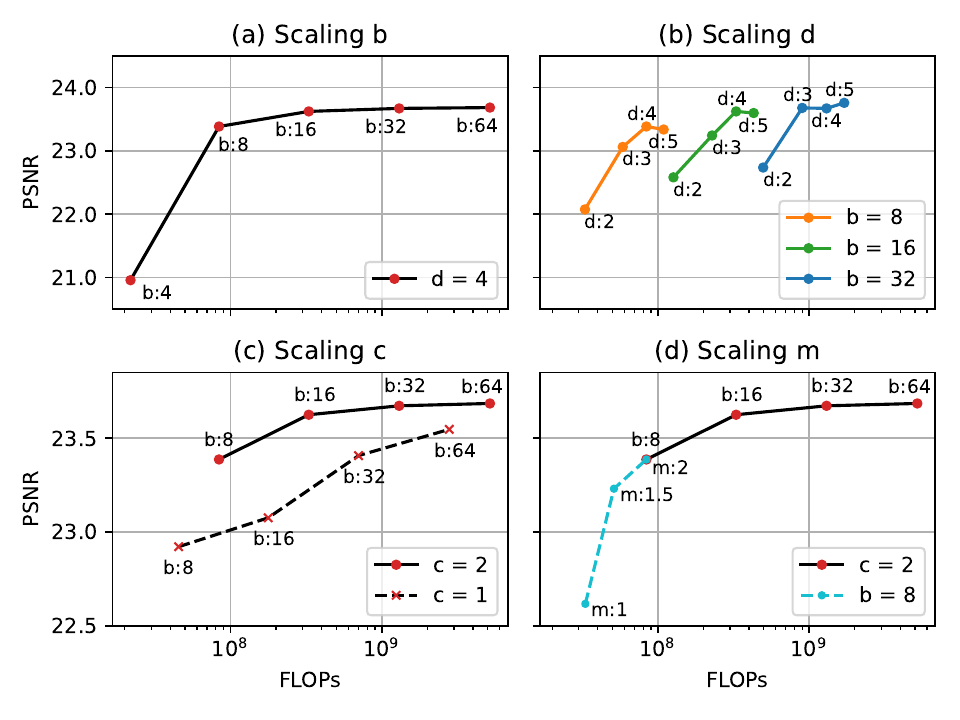}
    \caption{UNet hyperparameter (Figure \ref{fig:integrated}) scaling experiments. Shows how altering a specific hyper-parameter influences denoising performance and FLOPs. We only show PSNR results of \(\sigma\)=0.8, as the other results show the same trend. We find that \textit{b} and \textit{m} scale down efficiently, \textit{d} and \textit{c} do not.}
     \label{fig:denoising}
\end{figure}

\textbf{Exp. 1: The base feature map width \textit{b}.} In this experiment, we aim to find the relevant range of \textit{b}. Since \textit{b} is multiplied at every level, the number of feature maps throughout all layers depends on it, which makes it a powerful hyper-parameter. We use \textit{d} = 4 and the other hyper-parameters as in the original UNet, then we test \textit{b} \(\in\) \{4, 8, 16, 32, 64\}. 
The trend in Figure \ref{fig:denoising}{\color{black}.a} shows that the performance and FLOPs increase with \textit{b}. We observe that the trend is significantly disrupted by \textit{b} = 4.
Conversely, the performance difference between \textit{b} = 32 and \textit{b} = 64 is small, but the network size quadrupled. Therefore in further experiments, we focus on \textit{b} \(\in\) \{8, 16, 32\}.

\textbf{Exp. 2: The UNet depth \textit{d}.} Given the robustness of \textit{b}, we are interested in how reducing \textit{d} compares in terms of efficiency. Figure \ref{fig:denoising}{\color{black}.b} displays the performance of the architectures with the selected \textit{b} \(\in\) \{8, 16, 32\} testing \textit{d} \(\in\) \{2, 3, 4, 5\}. We observe that reducing \textit{d} causes a drop in denoising performance, whereas \textit{b} retains performance better, also in Figure \ref{fig:denoising}{\color{black}.a}. Therefore \textit{b} scales down more efficiently. The models with \textit{d} = 3 or 4 denoise most efficient. Especially for the smaller models, \textit{d} = 4 performs well.

\textbf{Exp. 3: The number of conv blocks per layer \textit{c}.} Does reducing \textit{c} further increase efficiency? To test this, we take the best-performing settings, \textit{d} = 4 and \textit{b} \(\in\) \{8, 16, 32, 64\}, and compare \textit{c} = 1 and \textit{c} = 2. Figure \ref{fig:denoising}{\color{black}.c} shows that the model with \textit{c} = 2 outperforms \textit{c} = 1. Therefore reducing \textit{c} does not benefit the model's efficiency.

\textbf{Exp. 4: The feature map multiplier \textit{m}.} We test if our smallest model could be further reduced in size by lowering \textit{m}. We take \textit{d} = 4 and \textit{b} = 8, and compare \textit{m} \(\in\) \{1, 1.5, 2\}. Figure \ref{fig:denoising}{\color{black}.d} shows that the reduction to \textit{m} = 1.5 retains performance. For \textit{m} = 1, the performance drops. Reducing \textit{m} to 1.5 could therefore be used to scale down the model when further reducing \textit{b} significantly decreases performance. 

\textbf{Conclusion.} To construct the reduced and baseline denoiser, we use the smallest and largest values from the found hyperparameter ranges. Resulting in baseline (UNet): \{\textit{b} = 32, \textit{d} = 4, \textit{m} = 2, \textit{c} = 2\} and reduced (UNet-S): \{\textit{b} = 8, \textit{d} = 4, \textit{m} = 1.5, \textit{c} = 2\}. Table \ref{tab:denoising} compares the two models for a selection of the noise levels. Although the reduced model has significantly fewer FLOPs and lower latency, the denoising performance is relatively similar to the baseline denoiser.

The UNet hyper-parameter experiments are replicated using MBConvs, which lead to similar findings. Moreover, the Conv UNet slightly outperforms the MB model. Therefore, the Conv model is used.

\begin{table}[]
\caption{Compares Baseline and Reduced UNet denoisers. The reduced model has significantly lower FLOPs and latency yet similar denoising performance.}
\centering
\resizebox{0.7\columnwidth}{!}{%
\begin{tabular}{@{}lccccccc@{}}
\toprule
\multirow{2}{*}{Model} & \multirow{2}{*}{FLOPs (M) \(\downarrow\)} & \multirow{2}{*}{Lat. (ms) \(\downarrow\)} & \multirow{2}{*}{Metric} & \multicolumn{4}{c}{Noise level   (\(\sigma\))} \\
 &  &  &  & 0.2 & 0.4 & 0.8 & 1 \\ \midrule
\multirow{2}{*}{UNet} & \multirow{2}{*}{1301.8} & \multirow{2}{*}{7.10} & PSNR \(\uparrow\) & 33.9 & 29.5 & 23.8 & 22.3 \\
 &  &  & SSIM \(\uparrow\) & 0.99 & 0.98 & 0.95 & 0.92 \\ 
\cmidrule(r){0-3}
\multirow{2}{*}{UNet-S} & \multirow{2}{*}{51.2} & \multirow{2}{*}{2.38} & PSNR \(\uparrow\) & 33.2 & 28.7 & 23.3 & 22.0 \\
 &  &  & SSIM \(\uparrow\) & 0.99 & 0.98 & 0.94 & 0.92 \\ 
\bottomrule
\end{tabular}%
}
\label{tab:denoising}
\end{table}

\section{Search space}
\label{app:search-space}
In Section \ref{subsec:NAS-classification}, different variations of the TF-NAS search space are used \cite{TF_NAS}. Table \ref{tab:searched-operators} displays the candidate operations and for which search space size they are used. The search space with 4 operators is constructed using the MBConvs without SE-layer, as this is most common in recent NAS methods \cite{EfficientNetV2, FB_net}. For the 6-operator search space, we add the possibility of using an SE layer on the operators where the kernel size is three and the expansion rate is three or six. We use the two smallest operators as they can be used for smaller target latencies too. The search space with 8 operators simply uses all combinations.

\begin{table}
\caption{Overview of the candidate blocks for the different search space sizes \{4, 6, 8\}. MBConv operators are used with different kernel sizes \textit{k}, expansion rate \textit{e}, and in- or excluding the squeeze- and excitation-layer.}
\centering
\resizebox{0.65\columnwidth}{!}{%
\begin{tabular}{@{}lcccccc@{}}
\toprule
Name & Kernel & Expansion rate & SE-layer & 4 & 6 & 8 \\ \midrule
MB-k3-e3 & 3 & 3 & - & \(\checkmark\) & \(\checkmark\) & \(\checkmark\) \\
MB-k3-e6 & 3 & 6 & - & \(\checkmark\) & \(\checkmark\) & \(\checkmark\) \\
MB-k5-e3 & 5 & 3 & - & \(\checkmark\) & \(\checkmark\) & \(\checkmark\) \\
MB-k5-e6 & 5 & 6 & - & \(\checkmark\) & \(\checkmark\) & \(\checkmark\) \\
MB-k3-e3-se & 3 & 3 & \(\checkmark\) & - & \(\checkmark\) & \(\checkmark\) \\
MB-k3-e6-se & 3 & 6 & \(\checkmark\) & - & \(\checkmark\) & \(\checkmark\) \\
MB-k5-e3-se & 5 & 3 & \(\checkmark\) & - & - & \(\checkmark\) \\
MB-k5-e6-se & 5 & 6 & \(\checkmark\) & - & - & \(\checkmark\) \\ \bottomrule
\end{tabular}%
}
\label{tab:searched-operators}
\end{table}

\section{Learned vs. Removed \texorpdfstring{$\beta$}{beta}: Additional results}
\label{app:nas-beta}
In Experiment 1 of Section \ref{subsec:NAS-classification}, we test the influence of removing \(\beta\) from the search approach. The models with Removed \(\beta\) significantly outperform the models with Learned \(\beta\) in accuracy. Besides, the found models are more similar for Removed than Fixed, Removed \(\beta\) differs only 0.04ms and 0.2\% accuracy, while Learned \(\beta\) differs 0.57ms and 1.4\% accuracy. This indicates that the search for Removed is more stable.

\begin{table}[h!]
\caption{Compares four searched models with target latency 6 ms. Trained on clean images. Two models are searched without \(\beta\) and the other two using learned \(\beta\). Removed outperforms learned \(\beta\).}
\label{tab:NAS-beta}
\centering
\resizebox{0.5\linewidth}{!}{
\begin{tabular}{@{}cccc@{}}
\toprule
Type & Search id & LAT (ms) \(\downarrow\)  & Acc (\%) \(\uparrow\) \\ \midrule
\multirow{2}{*}{Removed \(\beta\)} & 1 & 5.85 & \textbf{86.2} \\
 & 2 & 5.81 & \textbf{86.4} \\ \cmidrule(r){1-1}
\multirow{2}{*}{Learned \(\beta\)} & 1 & 5.04 & 84.9 \\
 & 2 & 4.47 & 83.5 \\ \bottomrule
\end{tabular}%
}
\end{table}